\documentclass[letterpaper, 10 pt, conference]{ieeeconf}
\IEEEoverridecommandlockouts{}
{}
 {}
\usepackage{hyperref}
\usepackage{url}
\usepackage{graphics} 
\usepackage{epsfig} 
\usepackage{mathptmx} 
\usepackage{times} 
\usepackage{booktabs}	 
\usepackage{amsfonts}	 
\usepackage{nicefrac}	 
\usepackage{microtype}	 
\usepackage{algorithm, multirow, xcolor}
\usepackage{algorithmicx}
\usepackage{algpseudocode}
\usepackage{mathtools}
\usepackage{graphicx}
\usepackage{cases}
\usepackage{array}
\usepackage{color}
\usepackage{float}
\usepackage{amssymb}
\usepackage{hhline}
\usepackage{cite}
\usepackage{makecell}
\usepackage{tablefootnote}
\usepackage{wrapfig}
\usepackage{amsmath}
\usepackage{amsthm, amssymb}
\theoremstyle{definition}

\usepackage{soul}
\usepackage{graphicx}
\usepackage{subfigure}
\usepackage{epstopdf}
\usepackage[skip=5pt]{caption}
\usepackage{dblfloatfix}

\title{\LARGE \bf
PA-LOCO: Learning Perturbation-Adaptive Locomotion for \\ Quadruped Robots
} 

\author{Zhiyuan Xiao, Xinyu Zhang, Xiang Zhou, and Qingrui Zhang
 \thanks{All authors are with School of Aeronautics and Astronautics, Sun Yat-sen University, Shenzhen 518107, P.R. China. Correspondence to: Qingrui Zhang ({\tt \small zhangqr9@mail.sysu.edu.cn})}
  }

\begin{document}

\maketitle
\thispagestyle{empty}
\pagestyle{empty}


\begin{abstract}
Numerous locomotion controllers have been designed based on Reinforcement Learning (RL) to facilitate blind quadrupedal locomotion traversing challenging terrains.  Nevertheless, locomotion control is still a challenging task for quadruped robots traversing diverse terrains amidst unforeseen disturbances. Recently, privileged learning has been employed to learn reliable and robust quadrupedal locomotion over various terrains based on a teacher-student architecture. However, its one-encoder structure is not adequate in addressing external force perturbations. The student policy would experience inevitable performance degradation due to the feature embedding discrepancy between the feature encoder of the teacher policy and the one of the student policy. Hence, this paper presents a privileged learning framework with multiple feature encoders and a residual policy network for robust and reliable quadruped locomotion subject to various external perturbations. The multi-encoder structure can decouple latent features from different privileged information, ultimately leading to enhanced performance of the learned policy in terms of robustness, stability, and reliability. The efficiency of the proposed feature encoding module is analyzed in depth using extensive simulation data. The introduction of the residual policy network helps mitigate the performance degradation experienced by the student policy that attempts to clone the behaviors of a teacher policy. The proposed framework is evaluated on a Unitree GO1 robot, showcasing its performance enhancement over the state-of-the-art privileged learning algorithm through extensive experiments conducted on diverse terrains. Ablation studies are conducted to illustrate the efficiency of the residual policy network. \footnote{Open source code is available at \href{https://github.com/QRZ-Lab/PA-LOCO}{PA-LOCO (IROS 2024)}}

\end{abstract}


\section{Introduction}

Model-free reinforcement learning method has demonstrated remarkable success in the advancement of locomotion controllers for legged robots \cite{tan_sim--real_2018, hwangbo_learning_2019, lee_learning_2020}. Previous research aimed to enhance the blind locomotion of legged robots on various complex terrains, such as steps, slopes, grass, mud, snow, and sand, maximizing their potential for outdoor operation. Due to the complexity and variability of the missions and working environments, quadruped robots are vulnerable to various unexpected perturbations or disturbances, such as collisions with dynamic obstacles or external sudden forces. Hence, it is critical to efficiently deal with unanticipated external force disturbances to endow quadruped robots with safe and reliable locomotion capabilities.

However, safe and reliable locomotion is a challenging task in the presence of unexpected perturbations, especially in case without any force sensors.  If left unaddressed, unforeseen perturbations, such as impact forces or sudden pushes, would propel a quadruped robot away from its stable locomotion, thus pushing the robot away from its intended trajectory. In severe scenarios, these perturbations can significantly deteriorate the locomotion stability of a quadruped robot, ultimately causing it to topple over. Therefore, it is beneficial, though challenging, for quadrupedal robots to actively compensate for such perturbations by using measurements from off-the-shelf onboard sensors, \emph{e.g.},  an inertial
measurement unit (IMU) and joint encoders \emph{etc}.

One straightforward idea to learn a robust locomotion control policy using reinforcement learning by the so-called domain randomization technique \cite{peng_sim--real_2018}. It involves training a policy on a variety of environments with randomized properties, such as perturbations with different magnitudes or sudden pushes with different directions. Via the domain randomization technique, a robot learning a robust policy that applies to different conditions. Such a policy is both passive and conservative. A more attractive solution is to allow robot react to external disturbances in an adaptive fashion using certain estimates based on onboard sensors.

\begin{figure}[!t]
 \centering
 \includegraphics[width = \linewidth]{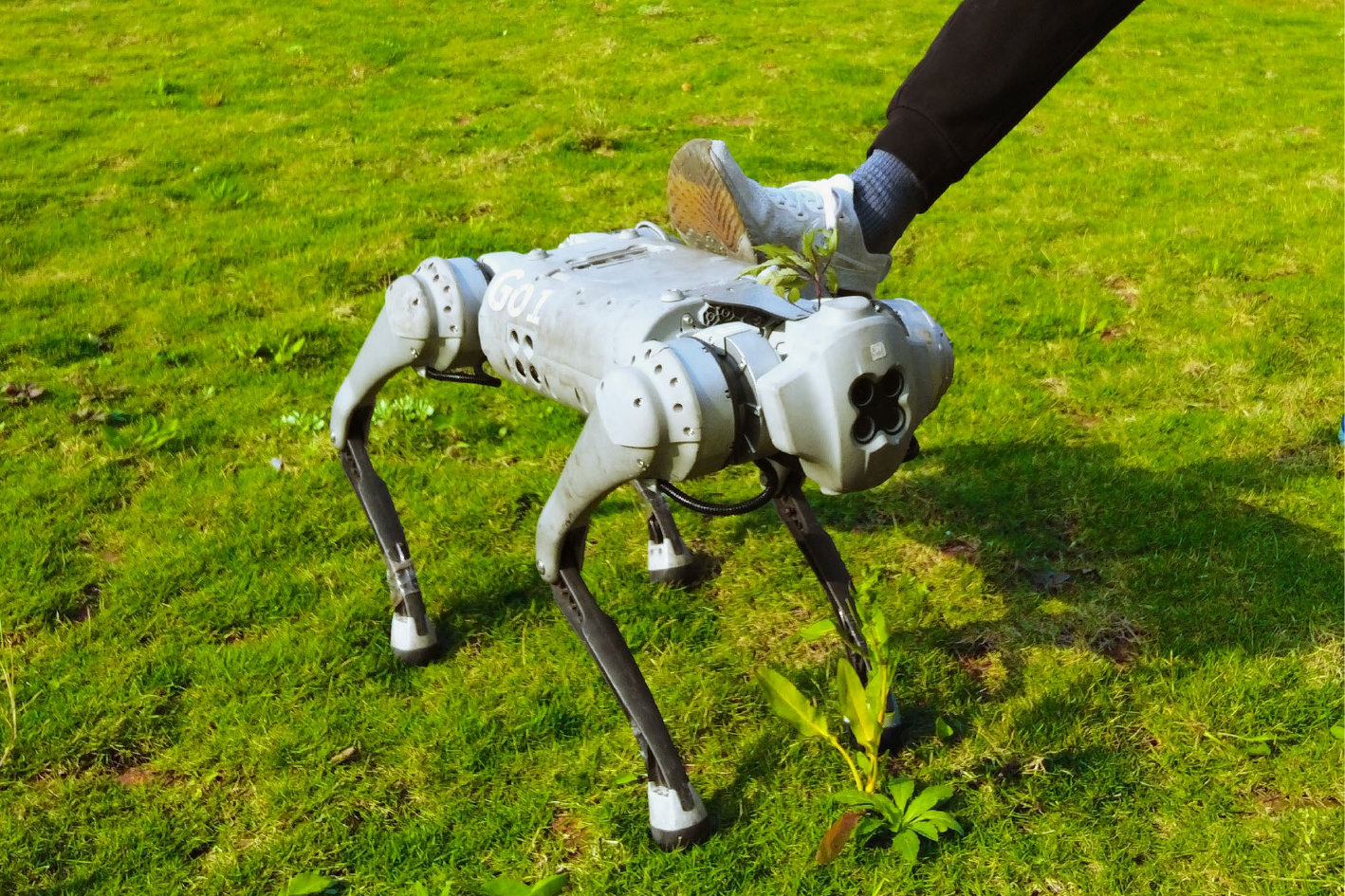}
 \caption{A Unitree Go1 quadruped robot is subject to a kick, while standing on a grass field.}
 \label{fig: snapshot}
\end{figure}

As an alternative, privileged learning provides a viable solution to learn a locomotion control policy that can actively handle external disturbances via certain estimates. In the privileged learning, a teacher-student structure is employed. The teacher policy is trained with additional or privileged information that is not available during testing or deployment. Such privileged information is embedded in a certain latent feature space. The latent feature space represents a lower-dimensional representation of the data that captures the underlying structure or patterns. A student policy is thereafter trained via supervised learning to imitate the behaviours of the teacher policy using available observations in real implementations \cite{lee_learning_2020, kumar_rma_2021, miki_learning_2022, kim_not_2023}. Privileged learning has been used to learn reliable and stable locomotion control policies for quadruped robots to traverse challenging terrains with the privileged information, such as terrain profiles, ground friction, and various robot states including trunk mass, velocity, and motor strength. However, the potential of the privileged learning method remains unexplored for the learning the a locomotion controller adaptable to external perturbations.

Furthermore, the student policy commonly suffers from performance degradation in comparison with the teacher policy in the privileged learning framework. The behaviour cloning process for student policy learning would naturally result in certain discrepancy between latent features inferred from available measurements and those from privileged information. Hence, the student policy is expected to be refined again. Another observation is that the one single encoder architecture in the existing privileged learning is not adequate enough for perturbation compensation. With the one encoder architecture, privileged perturbation information intertwines with other privileged signals. It is, therefore, impossible to distinguish the latent feature changes due to external perturbations from those by other variables (\emph{e.g.}, velocities or heading angle).

\begin{figure}[tbp]
 \centering
 \includegraphics[width = \linewidth]{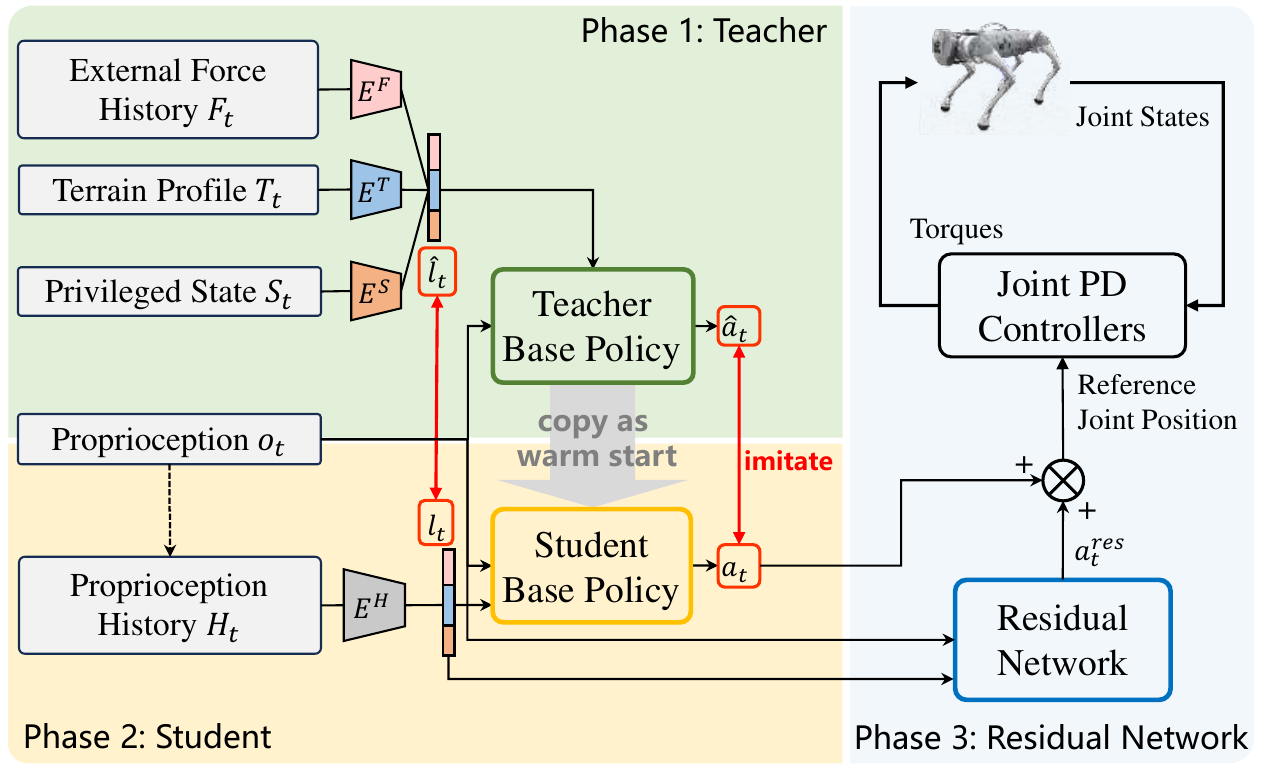}
\caption{The proposed PA-LOCO integrates a teacher-student framework with a residual network and multiple feature encoders. The training process involves three phases. In the first phase, the teacher policy is trained with proprioceptive observations $o_t$ and privileged information $F_t, T_t, S_t$ that is unknown for deployment. In the second phase, the student policy is trained using observations from the proprioceptive sensors. The student policy is learned to clone the teacher's actions and latent features by supervised learning. In the third phase, the residual policy network is trained to further enhance the performance of the student policy against perturbations.}
 \label{fig: framework}
\end{figure}

To tackle the aforementioned problems, this paper presents a teacher-student framework with multiple encoders as depicted in Fig. \ref{fig: framework}. The proposed framework aims to 1) achieve blind locomotion over diverse terrains; 2) actively compensate for external perturbations using existing onboard measurements. Through this framework, the learned policy is more robust against sudden force disturbances in comparison with the-state-of-the-art privilege learning algorithm. It also takes less time for the robot to recover its locomotion after the force impact. Experiments illustrate that the proposed framework can generate steady, adaptive, and robust locomotion in diverse perturbations and varied terrains. The overall contributions are three-fold:
\begin{enumerate}
    \item  A residual policy network is introduced to alleviate the student's performance degradation issue. The ablation study has shown that the residual policy network can improve the locomotion robustness and reduce the recovery time in the presence of disturbances.
    \item The privileged learning is improved by using a multi-encoder structure. With this modification, latent features from different privileged information are decoupled from each other, which reduce the potential mutual influence among different observations. Experiments have demonstrated that the multi-encoder structure is beneficial to the improvement of policy performance, \emph{e.g.,} robustness, stability, and reliability, \emph{etc}.
    \item The effectiveness of the latent feature embedding is analyzed sufficiently using simulated data. Extensive numerical simulations are performed to illustrate the effectiveness of the force encoder in distinguishing external forces of varying magnitudes and directions.
\end{enumerate}

The remainder of this paper is organized as follows. Section \ref{sec: Related Works} summarizes the related works about RL and privileged learning. Section \ref{sec: Reinforcement Learning} introduces the RL training method. In Section \ref{sec: Training}, the teacher-student architecture and more training details are provided. The experimental results are presented in Section \ref{sec: Results}. Conclusions are summarized in Section \ref{sec: Conclusion}.


\section{Related Works}\label{sec: Related Works}
\subsection{Reinforcement learning-based control}
In recent years, RL has been successfully implemented to deal with quadruped locomotion \cite{rudin_learning_2022}. The concurrently trained control policy and state estimator are capable of traversing slippery grounds and bumpy roads \cite{ji_concurrent_2022}. RL-trained networks also demonstrated steady and reliable locomotion capabilities on deformable terrains \cite{choi_learning_2023}. Some methods also achieve agile behaviors by combining RL with imitation learning to reduce the possibility of converging to local minima \cite{peng_learning_2020, jin_high_2022, escontrela_adversarial_2022, vollenweider_advanced_2023, li_versatile_2023, wu_learning_2023, li_learning_terrain_2023}. In addition, the learning-based method has been applied to other agile locomotion or manipulation skills, including fall recovery \cite{yang_multi_2020, hwangbo_learning_2019, yang_learning_2023}, back-flipping \cite{li_learning_2023, vezzi_two_2024}, jumping \cite{iscen_learning_2021, margolis_learning_2022, rudin_cat_2022, smith_learning_2023, vezzi_two_2024}, parkouring \cite{zhuang_robot_2023, cheng_extreme_2023}, rotating balls with limbs \cite{shi_circus_2021}, opening doors \cite{cheng_legs_2023}, and playing soccer \cite{ji_hierarchical_2022, ji_dribblebot_2023, huang_creating_2023}. In this paper, we focus on robust and adaptive quadruped locomotion control at various terrains in the presence of perturbations.

\subsection{Policy adaptation}
The domain randomization method is possible to allow the trained robust policy to be directly deployed to the robot without any modification. Peng \emph{et al.} achieved policy transfer for robust robotic arm operations via a domain randomization technique that randomly changes the parameters of the dynamic model in training \cite{peng_sim--real_2018}. Furthermore, Tan \emph{et al.} conducted an analytical evaluation of domain randomization efficacy within quadruped locomotion contexts \cite{tan_sim--real_2018}. However, it requires a trade-off between robustness and performance. To achieve adaptation to a new environment, Peng \emph{et al.} map the encoded dynamic parameters to a Gaussian distribution over a latent space\cite{peng_learning_2020}. However, the latent encoder needs to be re-trained in an offline fashion, when a robot is in a new environment with different settings. Recent advancements introduce online policy adaptation mechanisms, notably privileged learning and rapid motor adaptation (RMA).

\subsection{Teacher-student learning framework}
Privileged learning is a teacher-student learning framework, which is proposed by Lee \emph{et al.} for the locomotion control of quadruped robots \cite{lee_learning_2020}. The privileged learning encodes privileged terrain information, into a latent representation to solve the partial observability problem in blind quadruped locomotion \cite{lee_learning_2020}. It involves two phases of training. In the first phase, the teacher policy is trained using privileged information that is not available for deployment. In the second phase, the student policy is trained to replicate the teacher's behaviors using onboard sensor measurements that are easily accessible in real life. Kumar \emph{et al.} propose a comparable RMA framework enabling the real-time online policy adaptation to novel situations within fractions of a second \cite{kumar_rma_2021}. Moreover, the privileged learning framework is a specific case of policy distillation methods. A similar framework is employed where the teacher policy trained with privileged terrain information is distilled into a student with access to depth \cite{agarwal_legged_2023, cheng_extreme_2023}. Additionally, the teacher-student learning framework provides a viable solution to infer external privileged information using onboard sensor measurements \cite{zhang_terrain_2021, margolis_rapid_2022, fu_deep_2023}. Luo \emph{et al.} utilize the parameterized motor failure as privileged information to implicitly identify faults \cite{luo_ft_2023}. However, the existing algorithms are not satisfactory in training a perturbation-adaptive locomotion control policy. In this paper, the aforementioned issues will be addressed.


\section{Reinforcement Learning}
\label{sec: Reinforcement Learning}

Reinforcement Learning (RL) serves as a data-driven method that formulates the locomotion control problem within the framework of a Markov Decision Process (MDP). A parameterized control policy is learned through substantial trial and error using data from either simulation or real world. It should be noted that the stable blind quadrupedal locomotion control is a Partially Observable Markov Decision Process (POMDP) problem. To address the POMDP issue, privileged learning presents a promising framework to embed unavailable privileged information into a latent feature space that is implicitly inferred using proprioceptive sensor measurements in real-life deployment. With privileged learning, it is possible to develop a reliable policy for robust and steady quadruped locomotion  resilient against external disturbances.

\textbf{\emph{Observations:}}
The observation space of the teacher policy comprises proprioceptive sensor measurements, robot states, and external disturbances, as well as terrain profiles. The proprioceptive sensor measurements $o_{t}\in \mathbb{R}^{45}$ include trunk angular velocity $\omega_b$ obtained from the IMU, gravity unit vector in the body frame $\hat{g}$, joint positions $\{q_0, q_1,...,q_{11}\}$ and joint velocities $\{\dot{q}_0,\dot{q}_1,...,\dot{q}_{11}\}$ output by the joint encoder, high-level commands $\{v^{*}_{x},v^{*}_{y},\omega^{*}_{z}\}$, and the actions at the last timestep. The robot state information $S_{t}\in \mathbb{R}^{28}$ includes trunk linear velocity, trunk mass, the center of mass (COM), ground friction coefficient, foot contact forces with the ground, and contact states on the robot's thigh and calf. The time-series information $F_t\in \mathbb{R}^{30}$ represents the external force disturbances applied to the robot during the last 10 timesteps. The terrain profile $T_{t}\in \mathbb{R}^{187}$ consists of 187 height values sampled below the robot's trunk. The observation space of the student policy only contains proprioceptive sensor measurements $H_t$. The linear velocity command is scaled by 2.0. The angular rate command and trunk angular rates are scaled by 0.25. Joint velocities are scaled by 0.05.

\begin{table}[tbp]
\centering
 \caption{Reward functions and weights}
 \label{table: Reward}
\begin{tabular}{lcc}
\toprule
Name&Expression&Weight\\
\hline
Linear velocity tracking&$\exp(-\frac{||v_{b,xy}^* - v_{b,xy}||^2}{0.25})$ &1dt\\
Angular velocity tracking&$\exp(-\frac{||\omega_{b,z}^* - \omega_{b,z}||^2}{0.25})$&0.5dt\\

Linear velocity penalty&$v_{b,z}^2$ &-2dt\\
Angular velocity penalty&$||\omega_{b,xy}||^2$&-0.05dt\\
Trunk orientation&$||\hat{g}_{x}||^2+||\hat{g}_{y}||^2$&-1dt\\
Trunk height&$||h_b-h_b^*||^2$&-5dt\\
Joints acceleration&$||\frac{\dot{q}_{j-1}-\dot{q}_{j}}{dt}||^2$&$-1\times10^{-7}$dt\\
Joints torque&$||\tau_{j}||^2$&-0.0002dt\\
Action rate&$||q^*_{j-1}-q^*_{j}||^2$&-0.005dt\\
Self collision&$n_{collisions}$&-0.001dt\\
Foot air time&$\sum^4_{f=0}(t_{air,f}-0.5)$&1.0dt\\
Foot end position&$\sum^4_{n=1}\exp(-\frac{||p_{f}-p_{d}||^2}{0.02})$&0.3dt\\
\bottomrule
\end{tabular}
\end{table}

\textbf{\emph{Actions:}}
The action space contains 12-dimensional reference joint angles $a_{RL,t}$ for a quadruped robot. The position command signal $q^*_t$ for each joint is the sum of the default constant joint position $q_{default}$ and the RL output $a_{RL,t}$, so $q^*_t = q_{default} + a_{RL,t}$. The command signal $q^*_t$ is sent to low-level PD controllers with proportional and derivative gains are $K_p=20$ and $K_d=0.5$, respectively.

\begin{table}[bp]
\centering
\caption{PPO hyper-parameters}
 \label{table: PPO_paramters}
\begin{tabular}{cc}
\toprule
Parameter&Value\\
\hline
Learning rate&Adaptive\\
Batch size&98304 (4096$\times$24)\\
Mini-batch size&24576 (4096$\times$6)\\
Discount factor&0.99\\
GAE lambda&0.95\\
Desired KL-divergence&0.01\\
Entropy coefficient&0.01\\
Clip range&0.2\\
Number of epochs&5\\
\bottomrule
\end{tabular}
\end{table}

\textbf{\emph{Reward functions:}}
The total reward is a weighted sum of 12 terms shown in TABLE \ref{table: Reward}, consisting of task rewards and auxiliary rewards. The task rewards include linear and angular velocity tracking rewards. However, only optimizing the policy through task rewards leads to failure or weird motions. To encourage natural locomotion, auxiliary rewards are incorporated to foster the emergence of natural locomotion patterns. The trunk height and orientation rewards penalize the unsteady behaviors of the robot trunk. The rewards of joint acceleration and action rate penalize the dramatic change in the actual acceleration of the joint and the actions given. The joint torque reward encourages more energy-efficient locomotion. Self-collision term penalizes collisions between each joint and any vertical surface to encourage safe locomotion. The foot airtime term rewards the foot off the ground to ensure sufficient foot clearance during the swing. Following our previous work \cite{zhang_synloco_2023}, the foot end position term is added to penalize the deviation of the foot end position.

\textbf{\emph{RL policy:}} 
Proximal Probability Optimization (PPO) is selected to learn the feedback body controller in our design \cite{schulman_proximal_2017}. The hyperparameters of PPO are listed in TABLE \ref{table: PPO_paramters}.

\begin{table}[tbp]
\centering
\caption{Parameter setting for training}
\label{table: Domain Randomization}
\begin{tabular}{clc}
\toprule
&Parameter&Range\\
\hline
\multirow{4}{*}{\makecell{Randomized\\dynamics}}
&Trunk mass&[-1,1] kg\\
&COM displacement & [-0.03, 0.03] m\\
&Ground friction coefficient&[0.25, 1.5]\\

\multirow{5}{*}{\makecell{Trunk\\impulse}}
&External force magnitude $F_{x},F_{y}$&[-60, 60] N\\
&External force magnitude $F_{z}$&[-10, 10] N\\
&External force noise &[-2,2] N\\
&External wrench magnitude $\omega$&2.5 rad/s\\
&External wrench interval & 15 s\\

\multirow{4}{*}{\makecell{Sensor\\noises}}
&Trunk angular velocity noise&[-0.05, 0.05] rad/s\\
&Gravity vector noise&[-0.05, 0.05]\\
&Joint positions noise&[-0.01, 0.01] rad\\
&Joint velocities noise&[-0.075, 0.075] rad/s\\
\bottomrule
\end{tabular}
\end{table}

\textbf{\emph{Domain randomization:}}
Domain randomization techniques are employed in the training process to alleviate the sim-to-real gap issue. First, the dynamic parameters are randomized, including body mass, COM displacement, and ground friction coefficient, in each training episode to simulate the various environments. During each episode, a random force and wrench will be applied to the robot. Finally, we add noise to the sensor's feedback to increase the controller's robustness against measuring errors and sensor faults. The parameters used for domain randomization are shown in TABLE \ref{table: Domain Randomization}. All randomized parameters follow a uniform distribution.

\textbf{\emph{Physical Simulator and  Training Setup:}}
Isaac Gym environment is used to conduct our training \cite{makoviychuk_isaac_2021}. The quadruped robot is trained to follow high-level commands under random external disturbances on different terrains in parallel with 4096 agents. The simulation runs at 200 Hz, while the policy runs at 50 Hz. The maximum episode length is $20$ s (or $1000$ time steps). If the episode's duration reaches $20$ s  or the robot trunk collides with the ground, the episode is terminated and restarted. High-level commands include forward velocity $v^*_x$ ranging from [-1,1] m/s, lateral velocity $v^*_y$ ranging from [-1,1] m/s, and steering angular velocity $\omega^*_z$ ranging from [-1,1] rad/s. These commands are sampled from a uniform distribution every 10 seconds. At the beginning of each episode, the robot spawned in the air with the default pose and randomized high-level commands.


\section{Teacher-student Architecture}
\label{sec: Training}

\begin{table}[bp]
\centering
\caption{Network configurations}
\label{table: Network}
\begin{tabular}{clcc}
\toprule
Module (MLP)&Input&Hidden Layers&Output\\
\hline
$\pi^T$&$o_{t}, \hat{l}_{t}$&[512, 256, 128]&$\hat{a}_{t}$\\
$V$&$o_{t}, F_{t}, S_{t}, T_{t}$&[512, 256, 128]&$V_{t}$\\
$E^{F}$&$F_{t}$&[64, 32]&$\hat{l}^{F}_{t}$\\
$E^{T}$&$T_{t}$&[256, 128]&$\hat{l}^{T}_{t}$\\
$E^{S}$&$S_{t}$&[64, 32]&$\hat{l}^{S}_{t}$\\
$\pi^S$&$o_{t}, l_{t}$&[512, 256, 128]&$a_{t}$\\
$E^{H}$&$H_{t}$&[1024, 512, 256]&$l_{t}$\\
$R$&$o_{t}, l_{t}$&[256, 128, 64]&$a^{res}_{t}$\\
\bottomrule
\end{tabular}
\end{table}

\textbf{\emph{Teacher Policy Architecture:}}
The teacher policy consists of two types of MLP networks, namely the base policy network and the encoder networks. A three-layer Multi-Layer Perceptron (MLP) with ELU activation functions will parameterize the base policy $\pi^T$ and critic network $V$. The encoder networks include external force encoder $E^F$, terrain encoder $E^T$, and state encoder $E^S$, each of which has direct access to the corresponding privileged information. The external force encoder $E^F$ encodes the external forces history to the latent feature $\hat{l}^F_t$. The terrain encoder $E^T$ takes as input height value samples $T_t$ and then outputs $\hat{l}^T_t$. Privileged state information $S_t$ is encoded into the latent feature $\hat{l}^{S}_t$ by the state encoder $E^{S}$. The structure of each network is listed in TABLE \ref{table: Network}.

While it may appear simpler and more intuitive to use a single encoder to encode all privileged information, in real-world deployments, robots often experience issues with the coupled privileged information, particularly between external forces and other information. However, the simulation results do not reflect such an effect on motion. In practice, the robot struggles to maintain stable locomotion under rapidly changing velocity commands.

\textbf{\emph{Student Policy Architecture:}}
The student policy consists of a base policy network $\pi^S$ and a proprioception history encoder $E^H$. We choose to encode the feedback from the proprioceptive sensor collected in the last second,  i.e. the last 50 time steps. 

In the second phase, the student policy learns to adapt to different scenarios by imitating the teacher's actions $\hat{a}_t$ and latent features $\hat{l}_t=(\hat{l}^T_t, \hat{l}^S_t, \hat{l}^F_t)$. The student's base policy first copies the parameters of the teacher's base policy as a warm start. In addition, the training data is generated by rolling out the student's trajectories in simulation. The base policy and encoder are trained via supervised learning as in \eqref{eq:imi_loss}

\begin{equation}\label{eq:imi_loss}
    L = \left(\hat{a_t}-a_t\right)^2 + \left(\hat{l_t}-l_t\right)^2,
\end{equation}

\textbf{\emph{Residual Network Architecture:}}
The residual module consists of an MLP that takes the proprioception data $o_t$ and the latent feature $l_t$ as inputs and generates the residual signals to enhance the student's locomotion performance against perturbations. The residual policy network has a smaller size than the student policy as shown in Table \ref{table: Network}.  Throughout the training phase, the parameters of each encoder and the student's base policy are kept frozen. Analogous to the training process for the teacher policy, an asymmetric actor-critic architecture is employed, with the critic network initiated by the teacher's counterpart. The resultant RL signal is a weighted sum of the outputs of the student policy and the residual network with weights of $0.25$ and $0.1$, respectively.

\begin{figure*}[t]
 \centering
 \includegraphics[width = \linewidth]{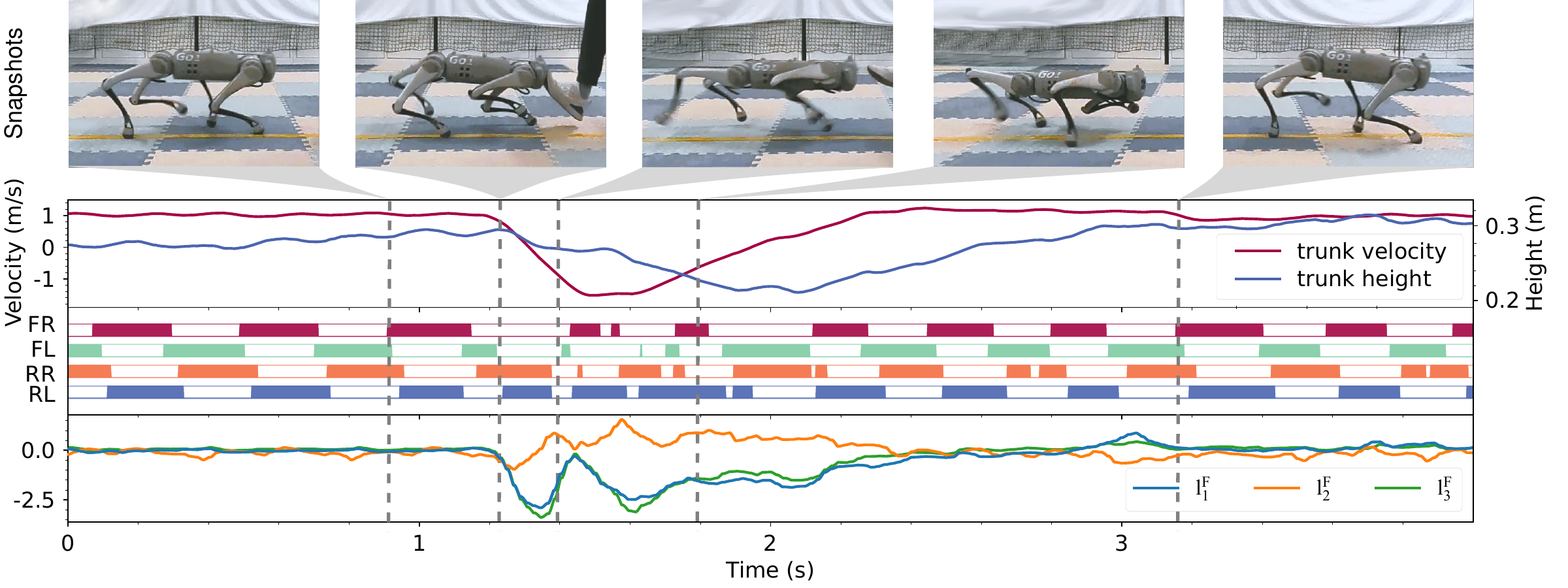}
 \caption{The locomotion behavior when subjected to external force impulses from the front. The trunk velocity and height responses are provided in the second image from the top. The feet's contact patterns with the ground (F/R denotes Front/Rear and R/L denotes Right/Left) are given in the third image from the top. The image at the bottom shows the plots of the force latent variables given by the force encoder.}
 \label{fig: Exp1}
\end{figure*}

\textbf{\emph{Curriculum Learning:}}
To learn adaptive behaviors under perturbation, a curriculum learning strategy similar to \cite{hwangbo_learning_2019} is introduced to steadily improve the locomotion performance via gradually increasing the perturbation during the training. If the mean tracking reward reaches a predefined threshold, the force magnitude will increase. Therefore, a fierce impulse is more likely to be applied to the robot. In addition, a terrain curriculum is also implemented to enhance the robot's ability to withstand perturbation on more complex terrain. At the beginning of the training process, the robot undergoes pre-training on flat terrain. Once the robot has demonstrated the ability to track with varying speed commands on flat terrain effectively, it advances to the second stage of training on complex terrain. Robots trained in parallel are then assigned to five types of terrains with a minimum level of difficulty: rough flats, smooth slopes, rough slopes, stairs, and discretized terrains. The robot only advances to more challenging terrain if it demonstrates exceptional velocity tracking performance in the last episode. If the velocity tracking performance falls below a certain threshold, the robot will return to the terrain with lower difficulty. The terrain difficulty will remain unchanged except in the aforementioned cases.


\section{Experimental Results} \label{sec: Results} 
Real-world experiments are conducted to validate the efficiency of the proposed PA-LOCO. The locomotion performance is evaluated by applying different forms of external perturbation to the robot or deploying the robot at diverse terrain configurations. Simulation experiments are performed to analyze  the efficiency of the latent representations of external forces with varying magnitudes and directions.

\subsection{Adaptive locomotion under external perturbations}

We kick the robot from its front to evaluate the locomotion adaptation and resilience under external perturbation as shown in Fig. \ref{fig: Exp1}. During the test, the robot is commanded to track a constant velocity of $1$ m/s. The OptiTrack motion capture system is used to obtain robot locomotion data for analysis.

Before the kick, the robot is able to track the constant velocity command, and consistently maintains a steady velocity of $1$ m/s. It trunk height would gradually reach to a value of $0.29$ as shown in Fig. \ref{fig: Exp1}. Moreover,  the force encoder outputs nill values and no noticeable variations are observed in its three components. 

Once kicked, the robot undergoes a noticeable deceleration, rapidly reducing its forward velocity of around $-1.5$ m/s in less than half a second due to the sudden front kick force as shown in Fig. \ref{fig: Exp1}. Correspondingly, the external force encoder can successfully capture these environmental changes, while the trunk height decreases to around $0.22$ m to enhance its own robustness. To tolerate instantaneous external force perturbations, the robot would like to lower its COM, thus increasing its foot contact frequency and adopting non-structured gait patterns to harness more locomotion stability. 

After the kick, the robot gradually resumes initial motion, while recovering a trot gait. Note that the robot first recovers to its initial speed ($1$ m/s), succeeded by a gradual convergence of the height to $0.3$ m with slight oscillations. Owing to the prominence of the velocity tracking reward over the trunk height penalty, the robot prioritizes velocity recovery. Ultimately, the external force encoder outputs $0$ again as the impact force disappears.

\subsection{Effects of force adaption and residual network}

We compare the performance of our framework with several state-of-the-art benchmark algorithms using a real-world quadruped robot (TABLE \ref{table: Metrics}). The comparison analysis is performed as the robot traverses on a plane at a velocity of $1$ m/s and are subject to a sudden lateral impact force midway. In addition to the proposed PA-LOCO, the following three benchmark algorithms are trained for comparison:
\begin{itemize}
    \item Robust: The policy is trained with domain randomization but without the force adaption mechanism $FA$ and a residual network $R$.
    \item SEFA: The policy is equipped with a single-encoder structure $SE$ and a force adaptation mechanism.  
    \item MEFA: It is an ablation study that removes the residual network and retains the multi-encoder $ME$ with a force adaptation module.
\end{itemize}

\begin{figure}[t]
 \centering
 \includegraphics[width = \linewidth]{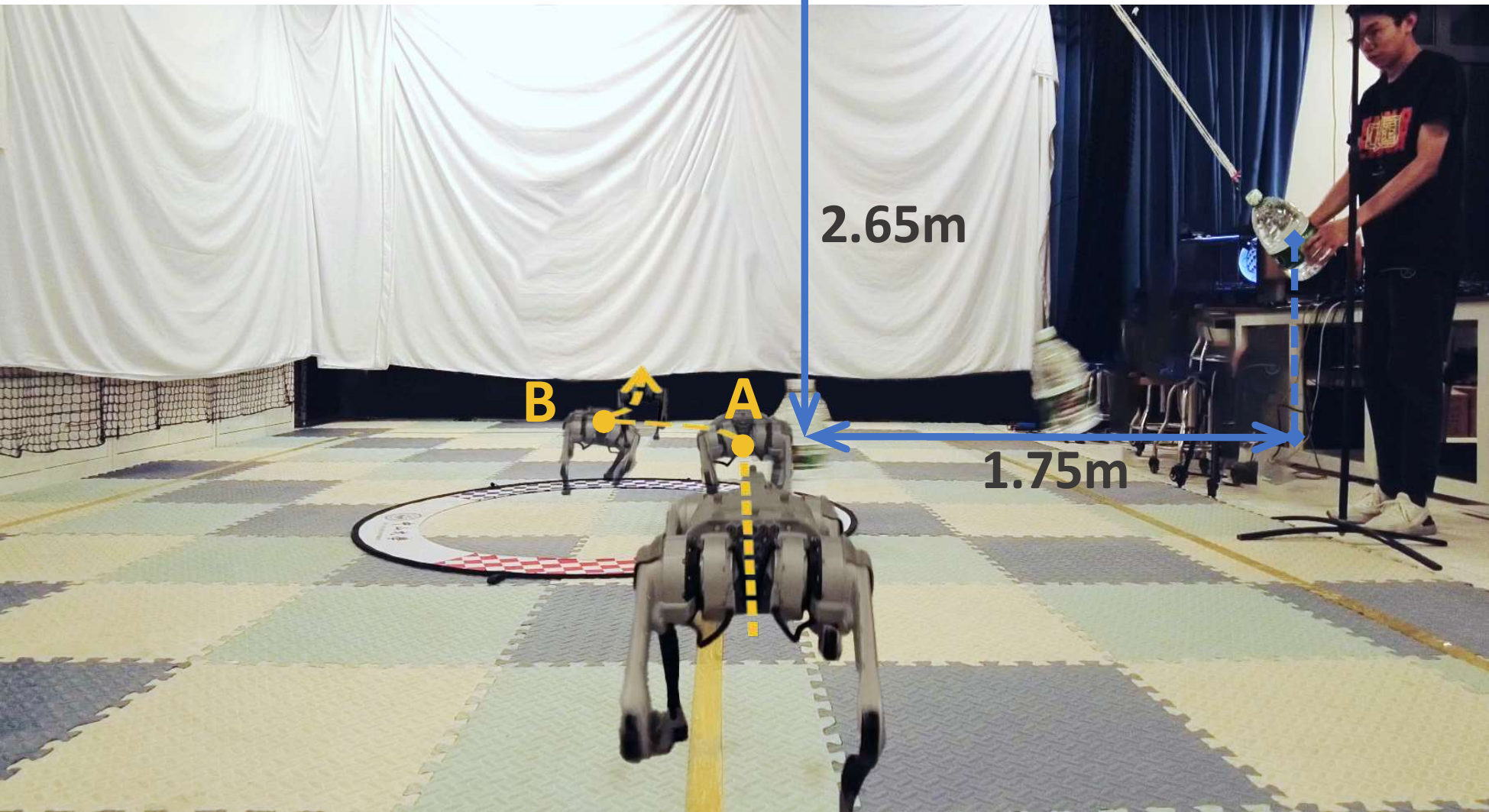}
 \caption{Indoor experiment setup.
 }
 \label{fig: indoor experiments}
\end{figure}

To generate a lateral impact force, a pendulum system is devised as shown in Fig. \ref{fig: indoor experiments}. This system comprises a suspended weight that swings around a pivot point on the top. The length of the pendulum is  $2.65$ meters. In each trial, the weight is lifted to a specific height such that the COM of the weight is $1.75$ m horizontally from the pivot, thereby maintaining a constant pendulum angle. When the weight reaches its lowest point, the COM of the weight is approximately $0.3$ m above the ground, which is equal to the height of the robot's trunk. At this point, the weight collides with the right side of the robot. The weights are chosen to be $2$ L and $4.5$ L bottled water, respectively, which have masses of $2.2$ kg and $4.6$ kg accordingly. These weights will generate sudden impacts in the robot's lateral, resulting in an instantaneous lateral velocity transition from $0$ to roughly $1.3$ m/s and $2$ m/s, respectively. 

\begin{table}[t]
\centering
\caption{Metric results for different algorithms}
\label{table: Metrics}
\resizebox{\linewidth}{!}{
\begin{tabular}{cccccc}
\toprule
 
 \multirow{2}{*}{Weights}&\multirow{2}{*}{Metrics}& \multicolumn{4}{c}{Algorithms}\\
 \cline{3-6}
&& Robust & SEFA & MEFA & PA-LOCO\\
\hline

\multirow{4}{*}{\makecell{2.2 kg}}&\multirow{4}{*}{\makecell{SR (\%)\\LO (m)\\RT (s)\\HO (m)}}& 90 (9/10) & 0 & $\mathbf{100}$ (10/10) & $\mathbf{100}$ (10/10) \\
&& 0.33 & - &  0.17 & $\mathbf{0.16}$ \\
&& 1.06 & - & 0.50 & $\mathbf{0.45}$ \\
&& 0.00 & - & -0.01 & $\mathbf{-0.02}$ \\
\hline

\multirow{4}{*}{\makecell{4.6 kg}}&\multirow{4}{*}{\makecell{SR (\%)\\LO (m)\\RT (s)\\HO (m)}}& 43 (3/7) & 0 & 80 (8/10) & $\mathbf{90}$ (9/10) \\
&& 1.46 & - & 0.64 & $\mathbf{0.59}$ \\
&& 1.98 & - & 0.89 & $\mathbf{0.75}$ \\
&& 0.00 & - & $\mathbf{-0.06}$ & -0.05\\

\bottomrule
\end{tabular}}
\end{table}

The following metrics are introduced to evaluate the performance of each algorithm (``A'' denotes the original lateral position (y axis in the body frame) at the moment before the impact, and ``B'' represents the maximum lateral position offset after the impact): (1) the success rate (SR) of surviving under lateral impact (No falling over), evaluating the robustness of the robot against external disturbances; (2) the average lateral COM offset (LO) of the robot due to the impact, evaluating the robustness of the robot against external disturbances; (3) the average recovery time (RT) after the impact, evaluating the robot's recovery capability after the perturbation; (4) the average trunk height offset (HO) after the impact, assessing the adaptation capability. 

The results of the average trunk height offset are listed in TABLE \ref{table: Metrics}. It can be observed that the robust policy can not adjust its trunk height in response to impact forces with varying magnitudes. Our PA-LOCO tends to lower the trunk height after impacts. In scenarios involving more severe impact forces by heavier weights, the force adaptation module can adjust the robot posture by lowering the COM even more. 

The results of the performance metrics are listed in TABLE \ref{table: Metrics}, which validate the efficiency of the force adaptation module and the add-on residual network. The performance of SEFA can not be measured since it exhibits highly unstable locomotion. The robust policy demonstrates limited adaptability to unexpected perturbations, resulting in greater lateral movements of the robot triggered by impulses, characterized by the metric LO. In contrast, both MEFA and PA-LOCO show lower values of LO and RT compared to the robust policy. It implies that the policies trained with an external force adaptation mechanism are more aware of unexpected perturbations. Furthermore, the decrease in both LO and RT values of PA-LOCO compared to MEFA also confirms PA-LOCO's ability to enhance the student's performance and expedite responses to lateral impacts.

\subsection{Analysis of the latent representation}

\begin{figure}[!ht]
 \centering
 \includegraphics[width = \linewidth]{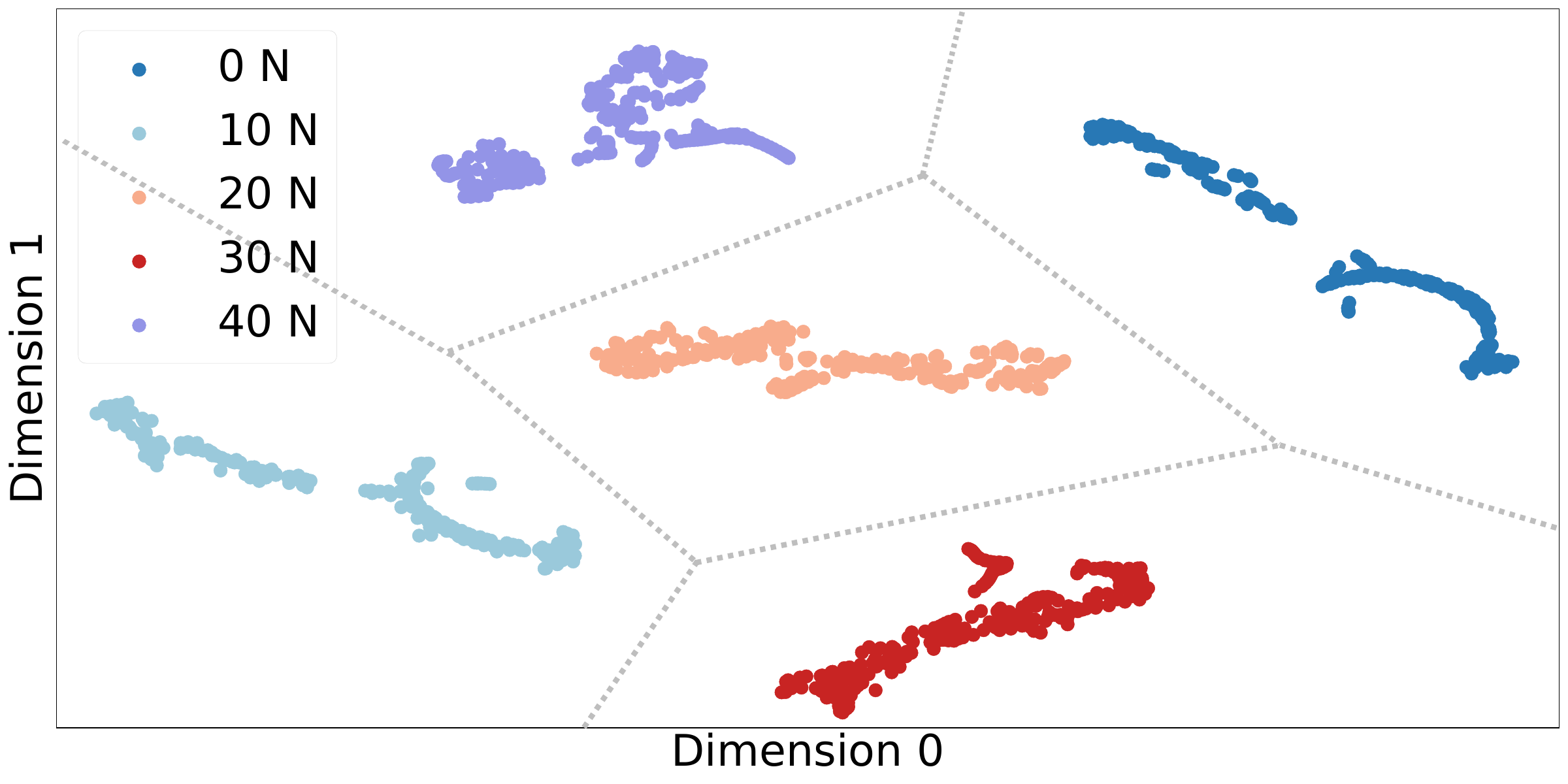}
 \caption{The t-SNE visualization of learned latent representation. In different trials, the robot is subjected to a constant backward force of different magnitudes when it moves forward. It indicates that the student policy is perturbation-aware due to the force encoder.
 }
 \label{fig: t-SNE}
\end{figure}

To gain deeper insights into the role of the encoder in discerning various perturbations of varying magnitude and directions, a series of simulation experiments are conducted. The first experiment entails the motion of a robot on flat terrain under a forward command of $0.5$ m/s while experiencing a constant backward force throughout the process. Across multiple trials, consistent external forces of varying magnitudes but uniform direction are applied, specifically $0$ N, $10$ N, $20$ N, $30$ N, and $40$ N, respectively. Multi-encoder is then used to encode the privileged terrain information $T_t$, state information $S_t$, and external perturbations information $F_t$ into a concatenated latent feature $l = (l^{T}, l^{S}, l^{F})$. Analysis based on the t-distributed stochastic neighbor embedding (t-SNE) plot, as depicted in Fig. \ref{fig: t-SNE} reveals the distinct distribution of latent features for different scenarios in the latent space. The analysis indicates that the learned policy is capable of discerning external forces of varying magnitudes, all in the same direction, through distinct latent features.

\begin{figure}[!ht]
 \centering
 \includegraphics[width = \linewidth]{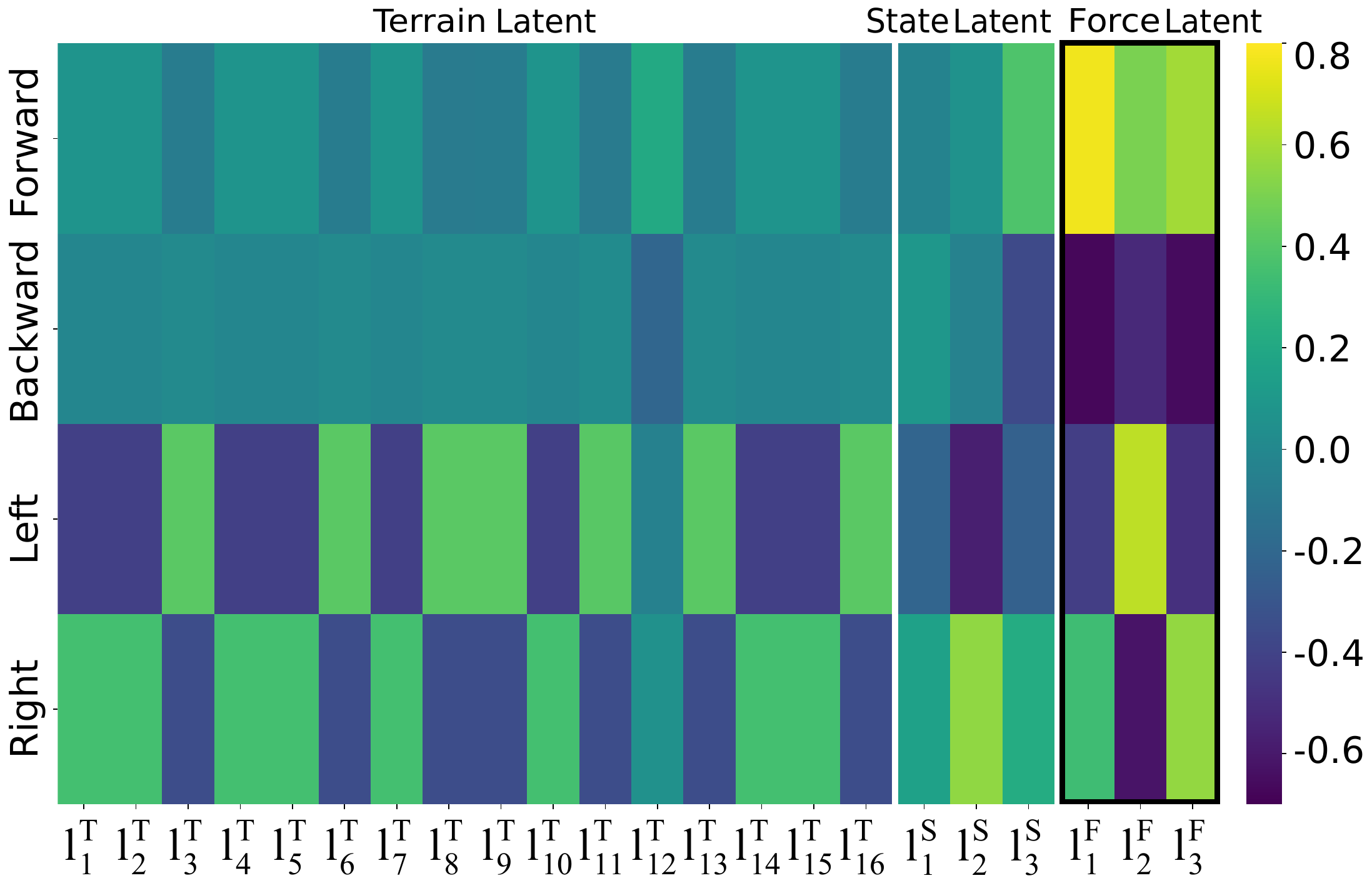}
 \caption{The correlation heatmap between latent features and different force directions. $l^{T}$, $l^{S}$, $l^{F}$ denote the latent features of the terrain profile $T_t$, robot state $S_t$, and external forces $F_t$ respectively. In different trials, the robot experiences a constant external force of $20$ N in different directions.
 }
 \label{fig: Heatmap}
\end{figure}

To further investigate the ability of a multi-encoder structure to discern consistent forces from different directions, the robot is given identical velocity commands on flat terrain while consistently experiencing external forces of $20$ N from four distinct directions: forward, backward, left, and right. The correlation heatmap between the force directions and each component of the latent feature is shown in Fig. \ref{fig: Heatmap}. It can be observed that considering each component of the latent feature of the external force $l^{F}$, there are four different forms of correlation between different directions and latent features of external forces. Note that the presence of forward forces is positively correlated with the three components of $l^{F}$, while the backward forces show a negative correlation. Similarly, for each component of $l^{F}$, different lateral forces show opposite correlations. Furthermore, some components of the state latent feature $l^{S}$ show a slightly higher correlation with the direction of the external forces. It is attributed to variations in the linear trunk velocity and foot-end contact state induced by external forces from different directions, similar to the robot's reaction after being subjected to a frontal kick illustrated in Session $A$. It implies that the multi-encoder structure enables the robot to distinguish between various types of privileged information, allowing it to make decisions based on the latent features.

\subsection{Locomotion control in various outdoor environments}

\begin{figure}[!ht]
 \centering
 \includegraphics[width = \linewidth]{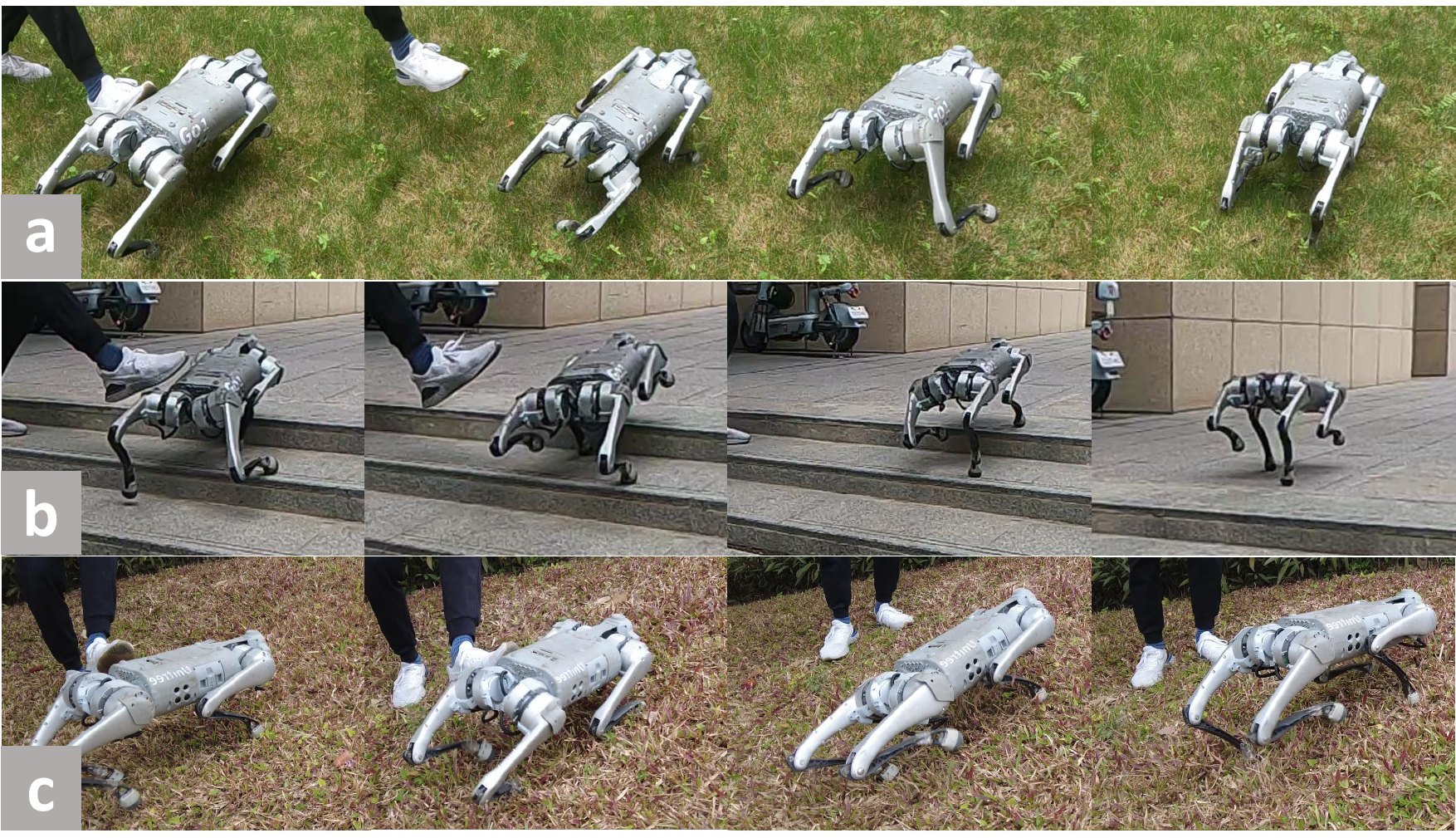}
 \caption{Outdoor experiments. (a) Grass. (b) Stairs. (c) Slope.
 }
 \label{fig: outdoor experiments}
\end{figure}

As shown in Fig. \ref{fig: outdoor experiments}, the proposed algorithm is further evaluated in diverse outdoor environments. Commands are sent using a joystick including the forward velocity that ranges from $[0,1]$ m/s. During the robot's movement, it will be impacted by a lateral kick. Notably, the robot demonstrates adaptability to unforeseen impulses across various terrains, encompassing grassy surfaces, staircases, and slopes. Following the cessation of perturbations, the robot gradually returns to its normal motion. It is noteworthy that when going uphill, specifically at a 20-degree incline as illustrated in Figure 6(c), the robot exhibits a noticeably reduced trunk height after encountering a sudden impact. Furthermore, a controller trained with one single-encoder architecture and a force adaptation mechanism is subsequently deployed on a real-world robot for testing within the aforementioned environments. However, this system exhibits excessive sensitivity to changes in the external environment and encounters challenges in maintaining stable locomotion, particularly at slightly elevated speeds. It can be seen in our supplementary video.


\section{Conclusions} \label{sec: Conclusion}

This paper presented a new teacher-student architecture with a residual policy and a multi-encoder structure for robust, reliable and steady locomotion control of quadruped robots. The residual policy with a smaller size is employed to mitigate the performance degradation issue when transferring a teacher policy to a student policy. The multi-encoder structure could decouple the latent features of external perturbations from those of other information, which can enhance the robustness, reliability, and responsiveness of the locomotion control. Diverse physical experiments were performed. It demonstrated that the learned controller could tolerate unexpected perturbations with sufficient robustness while traversing different terrains. Furthermore, the simulation results showed that the multi-encoder structure with an individual external force encoder can discern external forces of varying magnitudes and directions. However, the current design needs to weigh the importance of multiple encoder outputs according to different scenarios. Therefore, one future work will to introduce an attention mechanism to balance the outputs of multiple encoders in a more efficient way. 

\bibliography{references}
\bibliographystyle{IEEEtran}

\end{document}